# A Surgical Platform for Intracerebral Hemorrhage Robotic Evacuation (ASPIHRE): A Non-metallic MR-guided Concentric Tube Robot

Anthony L. Gunderman, Saikat Sengupta, Eleni Siampli, Dimitri Sigounas, Christopher Kellner, Chima Oluigbo, Karun Sharma, Isuru Godage, Kevin Cleary, Yue Chen

*Abstract*—**Intracerebral hemorrhage (ICH) is the deadliest stroke sub-type, with a one-month mortality rate as high as 52%. Due to the potential cortical disruption caused by craniotomy, conservative management ("watchful waiting") has historically been a common method of treatment. Minimally invasive evacuation has recently become an accepted method of treatment for patients with deep-seated hematoma 30-50 mL in volume, but proper visualization and tool dexterity remain constrained in conventional endoscopic approaches, particularly with larger hematoma volumes (> 50 mL). In this article we describe the development of ASPIHRE (A Surgical Platform for Intracerebral Hemorrhage Robotic Evacuation), the first-ever concentric tube robot that uses off-the-shelf plastic tubes for MR-guided ICH evacuation, improving tool dexterity and procedural visualization. The robot kinematics model is developed based on a calibration-based method and tube mechanics modeling, allowing the models to consider both variable curvature and torsional deflection. The MR-safe pneumatic motors are controlled using a variable gain PID algorithm producing a rotational accuracy of 0.317°±0.3°. The hardware and theoretical models are validated in a series of systematic bench-top and MRI experiments resulting in positional accuracy of the tube tip of 1.39±0.54 mm. Following validation of targeting accuracy, the robot's evacuation efficacy was tested in an MR-guided phantom clot evacuation experiment. The robot was able to evacuate an initially 38.36 mL clot in 5 minutes, leaving a residual hematoma of 8.14 mL, well below the 15 mL guideline suggesting good post-ICH evacuation clinical outcomes.**

This research is funded by NIH R01 NS116148. Corresponding Author: Yue Chen (yue.chen@bme.gatech.edu).

A. L. Gunderman and Y. Chen are with the Biomedical Engineering Department, Georgia Institute of Technology/Emory, Atlanta 30338 USA (e-mail: agunderman3@gatech.edu, yue.chen@bme.gatech.edu)

S. Sengupta is with the Vanderbilt University Institute of Imaging Science, Vanderbilt University Medical Center, Nashville, TN 37232 USA (e-mail: saikat.sengupta@vumc.org)

D. Sigounas is with The George Washington University School of Medicine and Health Sciences, Department of Neurosurgery, George Washington University, Washington, DC, US (e-mails: dsigounas@mfa.gwu.edu)

C. Kellner is with the Department of Neurosurgery, Icahn School of Medicine at Mount Sinai, New York, New York (e-mail: christopher.kellner@mountsinai.org)

I. Godage is with the Robotics and Medical Engineering (RoME) Laboratory, School of Computing, DePaul University, Chicago, IL 60604, USA. (e-mail: igodage@depaul.edu)

K. Sharma, E. Siampli, C. Oluigbo, and K. Cleary are with the Children's National Hospital, Washington, DC 20010 USA (e-mails: kvsharma@childrensnational.org, esiampli@childrensnational.org, coluigbo@childrensnational.org, kcleary@childrensnational.org

## I. INTRODUCTION

The World Health Organization (WHO) reported stroke as the cause of over six million deaths in 2019, second only to ischemic heart disease as the leading cause of death [1]. Of these deaths, intracerebral hemorrhage (ICH), which is the hemorrhaging of blood due to ruptured blood vessels within the brain [2], accounts for 20% of total stroke deaths, bestowing it the title of the deadliest stroke sub-type [3, 4]. With a one-month mortality rate as high as 52% [5, 6], the pathophysiologic mechanisms of ICH have been closely investigated to develop methods for reducing the mortality rate. Research indicates two distinct ways cerebral damage occurs. First, the collection of blood within the brain causes distortion and increased intracranial pressure, leading to mitochondrial injury and aberrant neurotransmitter release [3], both of which contribute to neuronal cell death and consequent cognitive dysfunction and decline [7, 8]. The second cause of damage is the release of the coagulant agent thrombin, a physiological response to the pooling blood [3]. Thrombin induces perihematomal edema (PHE), and the rate of PHE expansion is hypothesized to be a strong predictor of clinical outcomes after ICH [4]. Due to the aggressive nature of ICH, 40% of deaths occur within two days of symptom onset [3], suggesting a benefit to early hemorrhage evacuation following clinical decompensation to reduce intracranial pressure and PHE development rate [9]. However, despite the dismal outcomes of ICH and the need for urgency, "watchful waiting" or conservative management has been a common treatment method due to the cortical disruption and brain shift associated with large craniotomies in these usually critically ill patients [10-12] .

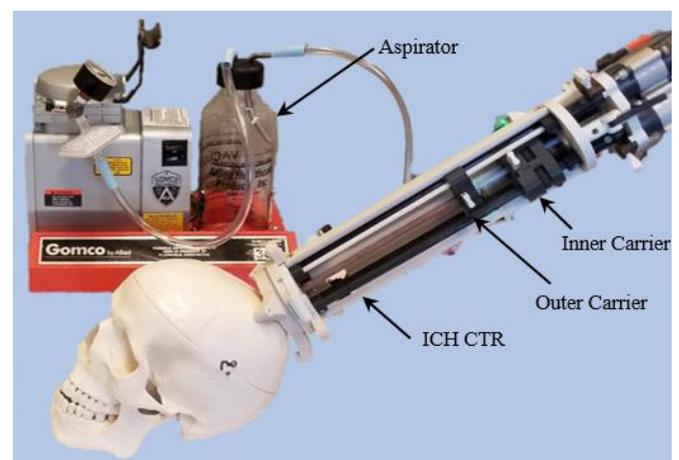

**Fig. 1.** The proposed MR-conditional Concentric Tube Robot (CTR) setup for ICH treatment.

Fortunately, in recent years, minimally invasive surgery (MIS) has proven to be a safer alternative to larger, open craniotomies for hemorrhagic clot evacuation and is an increasingly accepted method of treatment for patients with a deep-seated hematoma 30-50 mL in volume [9, 13, 14]. Compared to craniotomy, MIS utilizes a narrow surgical corridor, which has many benefits, including minimized disruption and scarring, reduced recovery time, reduced risk of infection, a lower mortality rate, fewer complications, reduced trauma, and lower medical costs [9]. MIS ICH evacuation can be classified as either passive or active [3, 6]. In the passive approach, a catheter is used to inject a thrombolytic agent at the site of the clot, allowing the clot to dissolve over multiple days [3]. In the active approach, the hematoma is removed in a short period of time through the use of aspiration systems with a ~4 mm ID evacuation tube [3]. In each case, the degree to which the blood is removed plays a critical role in the success of the procedure [6, 15-17].

The recently published Minimally Invasive Surgery Plus Alteplase for Intracerebral Hemorrhage Evacuation (MISTIE) III trial and a meta-analysis of all minimally invasive ICH evacuation randomized control trials suggest that near complete evacuation of a hemorrhage (< 15 mL residual hemorrhage) via a minimally invasive approach improve odds of a good clinical outcome [18]. Additionally, multiple clinical trials assessing the impact of endoscopic or microscope-assisted, minimally invasive evacuation of hemorrhages are ongoing (including INVEST (Minimally Invasive Endoscopic Surgery with Apollo in Patients with Brain Hemorrhage), MIND (A Prospective, Multicenter Study of Artemis, a Minimally Invasive Neuro Evacuation Device, in the Removal of Intracerebral Hemorrhage), and ENRICH (Early Minimally Invasive Removal of Intracerebral Hemorrhage)). The techniques evaluated in these studies have preliminarily demonstrated better clinical outcomes compared to conservative management [6].

The MIS techniques have also been validated in several commercial robotic studies. Wang et al. in 2019 used ROSA (Robotic Stereotactic Assistance, MedTech Surgical, Inc., USA) for spontaneous thalamic ICH (clot volumes of 5-15 mL) [19]. In that study, 35 patients were treated with catheter drainage using ROSA robot-assisted catheter deployment and pre-operative CT images, and 49 patients were treated with conservative management. The MIS procedure resulted in a lower Modified Rankin Score (measure of patient disability) compared to the conservative management approach, indicating an improved procedural outcome. Wang et al. in 2019 used Remebot (Remebot Technology Co., Beijing, China) to place catheters for initial hematoma (clot volumes of 20-40 mL) evacuation and thrombolytic infusion using pre-operative CT images [20]. The 3-month postoperative follow-up revealed improved neurological function, quality of life, and no mortality among all patients in the trial.

However, despite the many benefits and positive outcomes of MIS procedures, these procedures are still hindered by a lack of adequate hemorrhage visualization and tool dexterity and compactness [21]. Without real-time visualization, surgeons are unable to monitor the residual hemorrhage and brain shift, risking unintended crossing of the hemorrhage boundary and allowing suction to be applied to the normal parenchyma and vessels of the brain, potentially inducing re-hemorrhage [22]. Currently, to improve endoscopic visualization, infusion of saline or gas is used to permit direct visualization of the cavity; however, this has the deleterious effect of concurrently increasing intracranial pressure [23]. Additionally, endoscopic MIS procedures also have a limited workspace, which results in increased potential tissue disruption for evacuating hemorrhage volumes greater than 30-50 mL [24, 25]. This is due to the workspace constraint defined by the burr hole drilled into the skull, coupled with the limited degrees of freedom of current rigid endoscopes [26], requiring surgeons to tilt and torque the stiff metal neuroendoscope (i.e., Artemis™, BrainPath®) through the burr hole to achieve a wide workspace [27, 28]). This approach leads to brain tissue damage, creating what is colloquially referred to as a "cone of destruction" [28, 29]. Lastly, the large size of current devices (the diameter of commercial suction cannulas, the Artemis system, and the BrainPath system is 5-10 mm, 6 mm, and 11-13.5 mm, respectively) can lead to considerable tissue damage [6].

## II. Dexterous Needle-sized Robots For MIS

In recent studies, there has been a push to couple dexterous robotic interventional devices with intraprocedural visualization techniques as a way to ameliorate the limitations associated with conventional MIS. These dexterous, minimally invasive robots can generally be classified based on their method of achieving manipulation dexterity. These classifications are: (1) tendon-driven, (2) bevel-tipped, (3) shape memory alloy (SMA)-actuated, and (4) concentric tube robots (CTR).

Tendon-driven robots are a popular method of actuation in MIS due to their compact design and high level of distal dexterity [30-33]. These robots tend to be developed for a wide variety of clinical procedures. For example, Swaney et al. developed a flexural wrist to serve as a dexterous surgical tool for endoscopic procedures [34]. Kato et al. presented a tendon-driven continuum robot designed to fit within existing neuroendoscopes for the management of noncommunicating hydrocephalus and intracranial cysts, as well as in tumor biopsy procedures [35]. Chitalia et al. developed a meso-scale two-degree-of-freedom tendon-driven robotic endoscopic tool for the treatment of hydrocephalus using endoscopic third ventriculostomy. [36]. Gunderman et al. developed several tendon-driven tools for catheter-based brachytherapy and tissue biopsy [31, 32].

Bevel-tipped steerable needles are commonly used due to their design simplicity [37]. However, they are less commonly used in neurosurgery, specifically ICH, due to small evacuation diameters and the challenges in achieving tight curvatures. Additionally, bevel-tipped needles usually require multiple insertions to reach the target due to the difficulty in controlling the trajectory, risking extensive brain damage. Fortunately, Engh et al. have developed a bevel-tipped injection needle that uses a novel method for controlling the needle insertion trajectory. By leveraging the bevel tip and rotational degree of freedom of the needle, the trajectory is changed intraoperatively by continuously rotating the bevel-tipped needle at different percent duty cycles. This work enabled passive tumor reduction by circumventing the blood-brain barrier, allowing direct infusion of therapeutic agents for deep-seated brain tumors [38, 39].

Shape memory alloy actuators, though less common, are a low-cost, MRI-conditional actuation method with high power density. Kim et al. developed an example of an SMA-actuated robot for electrocauterization of brain tumors [40, 41]. Sheng et al. built an SMA prototype for ICH as well, developing torsional and bending joints that enable compact torsional and rotational motion in a meso-scale robot within the cerebrum, previously under-researched in SMA actuators [42].

The most successfully implemented MIS robotic device is the concentric tube robot. CTRs are commonly used due to their elegant modeling, simplistic design, and robust safety in terms of device durability [43]. Most importantly, the inner tube of CTRs can be directly used for aspiration purposes. Several groups have implemented CTRs for a variety of neurosurgical procedures. Bruns et al. have developed a CTR for orbital tumor resection that uses a chip-on endoscopic camera view [44]. Rox et al. developed a CTR with a compact differential drive for colloid cyst removal using an endoscopic camera [45]. Zhu et al. proposed a disposable, inexpensive CTR ICH robot costing less than $100 [46].

Despite these novel robotic approaches for improving dexterity within a confined workspace, none of the previously listed robotic systems completely addresses the lack of visualization inherent to MIS procedures. Several of the above listed approaches have only been tested in benchtop settings using external cameras [38-42, 46]. The remaining imaging modalities used in the aforementioned robotic systems (endoscopic and preoperative CT) do not provide the ability to account for intraoperative brain deformation and evacuation outcome monitoring, risking damage to the structures surrounding the hematoma. Recently, Godage, et al. have presented a concentric tube robot that allows for the first-time intraoperative image feedback for ICH evacuation using a CTR combined with CT images that are collected periodically during the hemorrhage removal [47, 48]. However, CT applies ionizing radiation to the patient, requiring a cost-benefit balance between dynamic visualization and ionizing radiation exposure [49]. In our recent work, we have used MR-guidance for a custom made CTR to provide intraprocedural information on the clot topology using periodic scans during the evacuation procedure [50, 51]. Although MRI provides superior imaging quality without ionizing radiation, its universal implementation for robotic intervention is impeded by the restriction of ferromagnetic materials, i.e., magnets, steel, iron, etc., as well as the reduction in imaging artifact and induced heating [52] associated with non-ferromagnetic metals, such as the nitinol tubes (see Section VI-D) and aluminum linear rails used by Chen et al. [50].

In this paper, we propose the design and evaluation of ASPIHRE, a skull mounted three degree of freedom MRI-conditional ICH evacuation robot made entirely of non-metallic materials (excluding optical encoders and optical limit switches), as shown in Fig. 1. This paper presents the first use of off-the-shelf plastic tubes, eliminating the artifact of the tube body, and discusses the corresponding analysis and validation of using it for ICH evacuation. This paper is arranged in the following way. Section III presents the design requirements for the robotic system. Section IV presents the robot hardware design, control algorithm, and surgical workflow. Section V presents the plastic tube selection and the corresponding kinematic modeling of the variable curvature shape. Section VI presents the experimental setup, procedure, and results. Section VII discusses the results. The paper is concluded in Section VIII.

## III. DESIGN REQUIREMENTS

### A. Clinical Requirements

As specified in the introduction, an evacuation tube with an inner diameter of ~4-5 mm is typically used for the evacuation of ICH hematomas [27-29]. Conventional endoscopes used for delivery of the evacuation tube are generally 10 mm in diameter [6, 53, 54]. Consequently, to maintain evacuation efficacy, the inner tube of our concentric tube robot must have an inner diameter of at least 4 mm; to prevent tissue disruption, the outer diameter of the outer tube must be 10 mm or less.

Several groups have demonstrated ICH evacuation using CTRs. These groups used evacuation tubes with a radius of curvature (RoC) ranging from 20 to 30 mm and lengths of curvature (LoC) ranging from 27 to 55 mm in the curved portion of the tube [48, 50, 55]. However, their tubes possessed an outer diameter < 2 mm. Due to the strain limitations of larger diameter tubes (see Section V-D), we elect to minimize the RoC of our tube based on the tube's material and its strain limit. Additionally, the translational distance requirement of the CTR platform is determined as 85 mm. This is based on the finding that the most common ICH occurs at the basal ganglia (close to the center of the cerebrum) [56, 57]. The average distance between the frontal bone, above the supraorbital foramen, and the basal ganglia is approximately 85 mm [58].

### B. Mounting Requirements

To ensure that a rigid body transformation relationship exists between the robot and the head, a skull-mounted stereotactic frame must be used to keep the robot fixed with respect to the patient. Additionally, the robot mounting must be ergonomic and allow fast attachment and detachment to reduce procedural preparation time.

### C. MR Imaging Modality Requirements

It is desirable to obtain intraoperative images of the hematoma during the evacuation procedure to monitor the treatment outcome. Thus, all actuators of the robot must be non-ferromagnetic, or electromagnetically decoupled from the MRI scanner, and all structures within the imaging volume must be non-metallic. There are different MR-conditional actuators that can be used in lieu of traditional electromagnetic actuators [59].

### D. Accuracy Requirements

Brain surgery is an inherently risky procedure and mitigating tissue disruption is a priority. Therefore, robot actuation/targeting accuracy is a priority. It should be noted that our robot possesses an evacuation tube with an inner diameter of ~4 mm, whereas previous CTRs had evacuation diameters ranging from 0.91 to 1.5 mm in diameter with tip positional root-mean-square-errors (RMSE) of 0.78 mm and 1.22 mm, respectively [50, 55]. Note that the ratio between accuracy and evacuation diameter is ~80% in prior studies. Thus, our aim in the preliminary study is to obtain a maximum RMSE of 3.2 mm (80% of 4 mm).

## IV. DESIGN, MODELING, AND IMPLEMENTATION

### A. Mechanical Design Overview

As mentioned above, the robotic system must mount rigidly to the patient's skull. However, due to anatomical variability between patients, tool material restrictions inside the MRI bore, and accessibility inside the MRI bore, the robotic system is broken up into three modular components. These include: (i) a custom-designed stereotactic frame (SF), (ii) a three degree of freedom concentric tube module (CTM), and (iii) a universal pneumatic transmission module (PTM), as shown in Fig. 2A.

The proposed system is mounted to the patient with three screws via a 3D-printed SF. The SF is printed out of polylactic acid (PLA) using a Stratasys F170 3D printer (F170, Stratasys,

MN, USA), providing print time as fast as 4 hours. This allows a custom skull mount to be made for a patient after the initial scan, ensuring robust mounting and reducing the time between clinical decompression and evacuation. The CTM is linked to the SF using a quarter-turn locking ring, as shown in Fig. 2A. This locking ring prevents rotations and translations of the CTM with respect to the SF. The quarter-turn functionality allows the ability to quickly exchange CTMs with different concentric tube parameters, permitting ICH evacuation for a variety of patients. The CTM is supported from the rear of the robot using two 6-DoF plastic support arms (Fig. 2B) manufactured using a stereolithography printer (Form 3B, Formlabs, MA, USA). The support arm was printed with Tough 2000 resin and supports a payload of 40 N when the arm is in a 90° configuration. Note that the CTM can be removed completely, allowing access for different tools, while the SF remains mounted to the skull, serving as a guide to define the tool insertion pose. The PTM is coupled to the CTM through a series of plastic locating dowel pins, snap-fit latches, and splined transmission shafts. The splined transmission shafts are critical for the modular robotic approach as they ensure the PTM can align with the CTM given any actuator rotation angle. The proposed mounting setup can be seen in Fig. 2B. When aspirating, the inner tube is connected to an aspirator (Gomco Model 300, Allied Healthcare, FL., USA) through a pneumatic union. The aspirator evacuation pressure, ranging from 0-30 mmHg, is regulated through the use of a thumb-screw valve. The maximum pressure is adjusted based on imaging feedback of the hematoma evacuation procedure. For example, if the aspirator is engaged, but the hematoma is unchanging, the aspiration pressure can be increased to induce evacuation flow.

### B. Concentric Tube Module (CTM)

As previously stated, the robot must be devoid of metallic materials to prevent image quality reduction during the intraoperative MR-imaging protocol. Thus, the CTM is made entirely of non-metallic components. The CTM chassis, inner and outer concentric tube carriers, and the inner concentric tube gears are all 3D-printed out of PLA. The CTM is 66 mm in diameter and 242 mm in length, providing a total insertion depth of 156 mm, which is able to reach any ICH target. The concentric tube carriers are translated along carbon fiber rods (OD: 6.35 mm) with Delrin® bushings using plastic translational lead-screws consisting of off-the-shelf 10-32 all-thread made of acetal and 10-32 polycarbonate hex nuts. These two plastic materials were chosen due to their low frictional coefficient and high stiffness.

The outer concentric tube carrier retains a fiberglass outer tube (OD: 7.93 mm, ID: 6.2 mm) using a polygon spline and retaining clip, as shown in Fig. 3A. The polygon spline is used to prevent radial motion while the retaining clip is used to prevent axial motion. The male part of the polygon spline is fixed to the tube body using epoxy (9200 Structural Epoxy, MG Chemicals, BC, Canada). The female part of the spline for the outer tube is embedded into the outer tube's carrier, as shown in Fig. 3A. In a similar manner, the inner concentric tube carrier retains a nylon inner tube (OD: 6 mm, ID: 4 mm). However, the female part of the polygon spline for the inner tube is embedded into a spur gear. The spur gear is held in place by plastic radial ball bearings (B6800B3G, igus®, RI, USA), allowing rotation when applied by the pinion driven by the PTM, as shown in Fig. 3B. The carriers are homed using optical limit switches (2HW144083OG5A, MakerHawk, USA) and the rotational motion of the inner tube is homed using the index pin on a quadrature digital encoder.

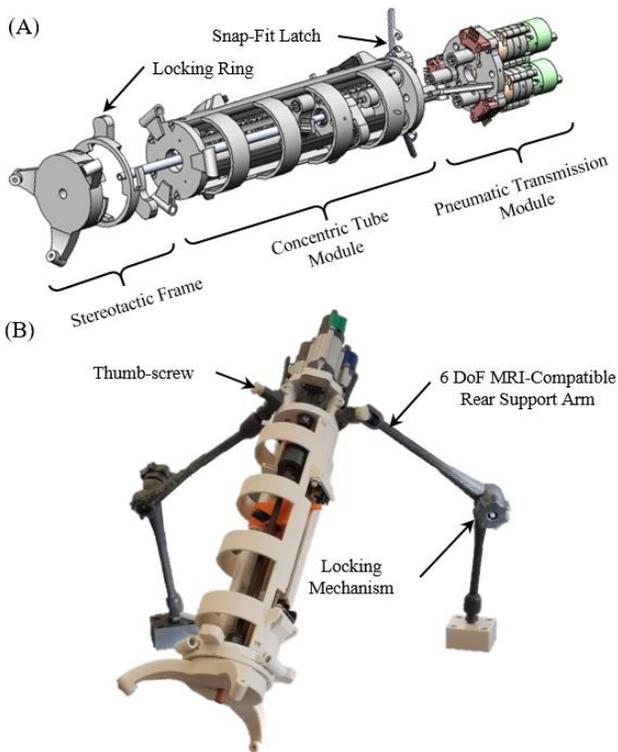

**Fig. 2.** (A) The ICH CTR can be seen in its disassembled and (B) assembled state. The front end of the robot is supported with three screws using the SF, whereas the rear part of the robot is supported using two plastic 6-DoF MRI-safe support arms.

## C. Pneumatic Transmission Module (PTM)

The PTM consists of three pneumatic bi-directional radial inflow turbines (OD: 28 mm), which were modified from our prior work [60] with reduced dimensions. These motors are attached to a modular plastic planetary gearbox (Tamiya 72001) and are coupled to their corresponding lead screw through splined 3D-printed shaft couplings (Fig. 4). Each motor is driven by two 10-meter pneumatic lines routed through the MRI waveguide allowing bi-directional rotation. It should be noted that the 10-meter pneumatic lines are a unique feature of this motor, allowing decoupling of the electromagnetic interaction between air distribution (valves, electronics, and compressor) and the MRI. It is generally seen that other pneumatic actuators require pneumatic tubing less than three meters, necessitating the use of Faraday's cages in the MRI room to eliminate image artifact or safety concerns [59].

## D. Mechatronic Hardware and Control Method

The pneumatic motors are controlled using a 16-bit DMC-4163 Galil Motion Controller (DMC-4163, Galil, CA, USA), which measures absolute position in terms of encoder pulses using a quadrature digital encoder and disk (EM1-1-1000-I and DISK-1-1000-375-IE, US Digital, WA., USA). The flow to the motors is regulated using 5/3 normally closed pneumatic flow control valves (MPYE-5-M5–010-B, Festo, US). The 5/3valves require a 0-10 V signal, with 5 V representing normally closed. However, the Galil controller provides a signal output of -10-10 V. Thus, a differential amplifier was designed to convert the -10-10 V control command to 0-10 V. The DMC-4163 low-level control normally operates with user-defined PID values. However, our system has two contradictory complexities that preclude the implementation of standard PID.

1. Our turbine motors have a high nominal speed (11,000 rpm). Consequently, a large gear reduction must be used (>100:1). The frictional forces induced by the gear reduction cause a significant dead-band region when the proportional gain is too low, causing the motor to remain stationary when the motor initiates movement.

2. Our imaging modality requires the motor to use 10-meter pneumatic lines, which introduces a significant delay of ~0.352s. Consequently, steady state oscillation is exhibited when the motor uses a high proportional gain.

In order for the motor to operate efficiently and maintain the accuracy required by the clinical requirements, a control law was developed and implemented that modifies the proportional gain of the PID controller as a function of the error (continuously) and the state of the motor (i.e., not moving vs. moving). This controller operates in three regions: Region I, when the error is below a specified error tolerance (1°), all gains of the PID controller are set to zero; Region II, when the error is greater than the specified tolerance (1°), but is less than a higher specified error threshold (90°), the proportional gain changes as a function of error and the motor state; Region III, when the error is greater than the specified error threshold (90°) the proportional gain remains constant. In Region II and III, the integral gain and derivative gain are always 0 and 4000, respectively.

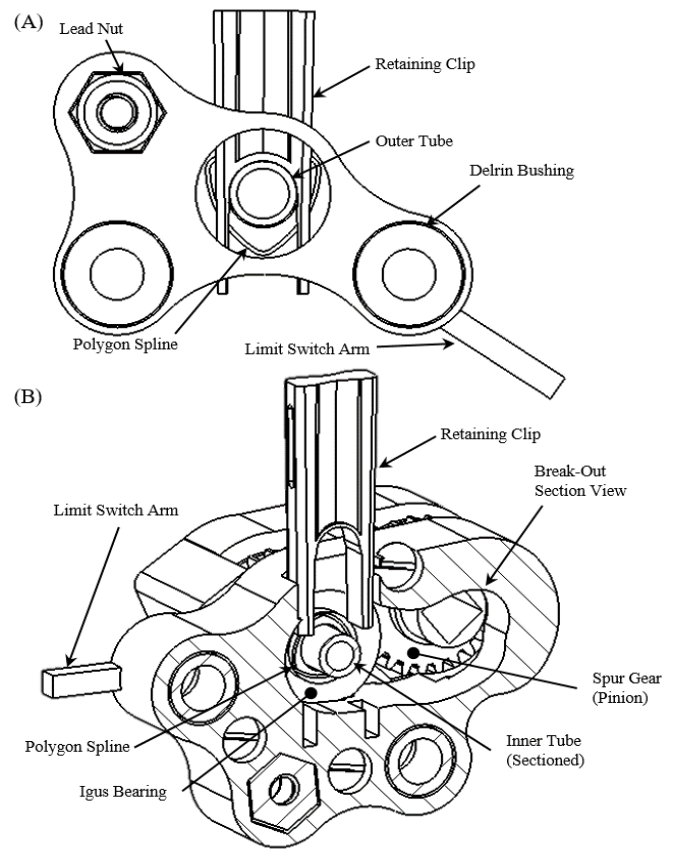

**Fig. 3.** (A) The outer carrier chassis can be seen holding the outer tube with the polygon spline and retaining clip. (B) In a similar fashion, the inner carrier can be seen holding the inner tube. Note that this view is sectioned to provide visual details of the retention method using the retaining clip. The pinion gear (depicted above) is driven by the PTM through a square Delrin® shaft. It is meshed with the spur gear of the inner tube, which rotates the inner tube.

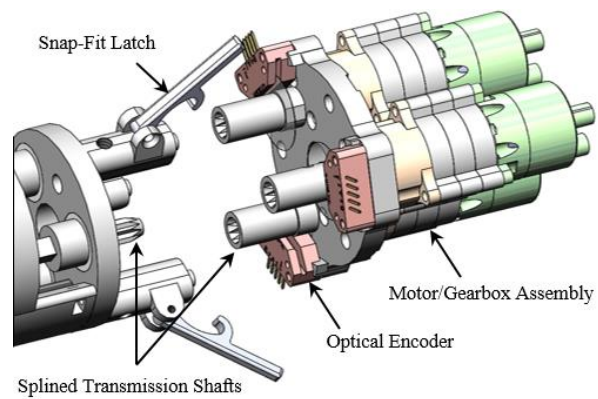

**Fig. 4.** The PTM can be seen decoupled from the CTM. Note the splined shafts which permit axial translation for decoupling, while radial play is eliminated. The encoder disks are held in place by the splined transmission shafts.

The function of the proportional gain in Region II begins by considering the equation for the voltage signal produced by the Galil controller given an arbitrary proportional gain and all other gains equal to zero, as shown:

$$V = K_p * err * \frac{V_{range}}{2^n} \tag{1}$$

where $V$ is the voltage signal produced by the Galil controller in the -10-10 V range, $err$ is the error in pulse counts, $V_{range}$ is the full-scale range of the voltage (20 V), and $n$ is the number of bits of the controller (16 bit). Due to the dead-band region in region II, there is a minimum required voltage value needed to initiate or continue movement, which will be measured experimentally. Therefore, we can rewrite (1) to solve for the proportional gain in Region II, providing the full control law as described below:

$$K_P(err, \omega) = \qquad\qquad\qquad\qquad Region$$

$$\begin{cases} 0 & |err| \le \delta & \text{I} \\ \frac{V_{start}}{err*\frac{20}{2^{16}}} & \delta < |err| \le err_{thresh} \text{ and } \omega = 0 & \text{II} \\ \frac{V_{d-b}}{err*\frac{20}{2^{16}}} & \delta < |err| \le err_{thresh} \text{ and } \omega \ne 0 & \text{II} \\ K_{P,const} & |err| > err_{thresh} & \text{III} \end{cases} \tag{2}$$

where $K_p$ is the proportional gain of the controller as a function of the error and rotational speed $\omega$, $\delta$ is the acceptable error tolerance of the motor, $err_{thresh}$ is the upper error bound of region II, $V_{start}$ is the minimum voltage signal necessary to move the motor when the rotational velocity is initially zero in region II, $V_{d-b}$ (typically lower than $V_{start}$) is the minimum voltage signal necessary to avoid the dead-band region as the motor approaches the target in region II with a non-zero rotational velocity, and $K_{p,const}$ is the constant proportional gain assigned when in region III. Note that the corresponding region is listed to the right of each equality in (2). Fig. 5 presents a plot of the corresponding voltage signal as the motor moves from an error higher than $err_{thresh}$ (90°) to an error lower than $\delta$ (1°). The corresponding block diagram of the control system can be seen in Fig. 6.

Note that for both axes, it is observed that the start voltage $V_{start}$ is significantly higher than the dead-band voltage $V_{d-b}$. This is possibly due to a higher coefficient of static friction as compared to the coefficient of kinetic friction of the system.

### E. Intraoperative MR Imaging

The MR-guided intervention was performed in a 70 cm wide bore, Philips 3 Tesla Ingenia Elition MRI System with a 2-channel body transmit/2-channel flex coil receive setup. The robot and clot were localized using 3D scout imaging and MR visible fiducials placed on the SF. The robot is registered using rigid-body registration with respect to the SF's fiducials. Dynamic images were taken of the region of interest (ROI) using single slice spoiled gradient-echo imaging along the plane intersecting with the plane the deflectable portion of the inner tube lies in. Single slice imaging was performed using a field of view of: 240×240 mm², 1×1 mm² resolution, TR/TE = 5/2 ms, and a dynamic scan time of 1 s/image. Prior to the robot performing a move to a target, the inverse kinematics are used to calculate the rotational angle of the inner tube in the robotframe and the slice plane is set to correspond to this angle in the robot's frame using point-based registration methods [61]. Imaging is then initiated in synchrony with the start of the clinical/experimental procedure. When a new target is desired, the slice plane is adjusted based on the inverse kinematics and the robot is then allowed to move.

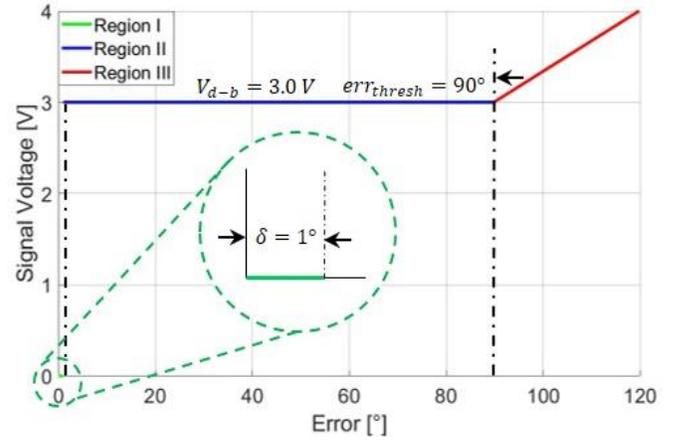

**Fig. 5.** The schematic diagram of the voltage signal of a motor moving from high error to low error (right to left) with $V_{d-b} = 3$ V, $K_{P,const} = 9.8$. $\delta = 1°$ (11 pulses), and $err_{thresh} = 90°$ (1000 pulses). In region III, the error is significantly large such that a constant proportional gain is able to move the motor towards the setpoint. In region II, the gain is a function of error, ensuring that the voltage is high enough to avoid the dead-band region, but low enough to prevent oscillatory behavior. In region I, the tolerance is reached and the voltage is set to zero.

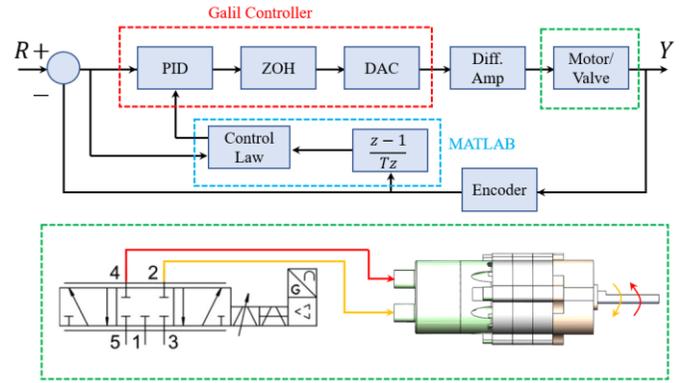

**Fig. 6.** The block diagram operates under a control law defined by the rotational velocity of the motor and the error. The input is provided by the surgeon's prescribed target point. The proportional gain is assigned to the Galil controller (red) by MATLAB (blue) using the relationship defined by equation (4). The integral gain is set to zero and the derivative gain is set to 4000, the maximum derivative gain available by the Galil controller. The resulting command signal from the Galil controller is converted to the voltage range of the pneumatic flow control valve using the differential amplifier. The motor is then moved clockwise (orange) when the command signal is > 5 V and counter-clockwise (red) when the command signal is < 5 V.

### F. System Control Diagram and Evacuation Workflow

In our envisioned clinical workflow, a preoperative scan would first be performed to localize the clot. This would provide the clinician with the anatomical map used to determine the optimal insertion path of the inner/outer tube pair, defining the geometry of the SF. Using this geometry, the SF would be manufactured and mounted to the skull such that the desired entry pose between the hematoma and burr hole location is aligned with the robot's optimal insertion vector. After mounting, the clinician would drill a burr hole and the patient would be transferred to the MRI. Fig. 7 shows the envisioned operational setup inside the MRI control room and MRI-scanner. The controller, pneumatic regulators, compressor, aspirator, and work stations (MRI and robot work stations) are separated from the MRI scanner through the wave-guide and remain in the control room, where the clinician resides. The robot is connected to the flow regulators and aspirator with 10-meter pneumatic lines; the robot is connected to the controller with 10-meter shielded cables linking the optical encoders and limit switches to the controller. The clinician closes the feedback loop between the robot's tip location and target locations in the clot through intraoperative imaging feedback. Due to the MR-conditionality of the robot, the robot can be setup while the SF is being printed, allowing a parallel setup sequence, negating additional lead-time prior to the procedure.

The in-scanner work-flow can be seen in Fig. 8. The robot is first connected to the robot's workstation and the Galil Motion Controller through the Graphical User Interface (GUI) and homed (~5 minutes). After homing, the robot is mounted to the SF from the superior end of the MRI bore and registered to the MRI's coordinate system using 3D scout imaging and rigid-body registration with respect to the SF's fiducials (~5 minutes). The clot is also localized in the prior step. Following localization, the slice plane is set to coincide with the inner tube's rotational position and intraoperative imaging is initiated, informing the clinician of the clot topology and the projected path of the CTR on the 3D scout image using intraoperative image feedback, closing the feedback loop between the robot and the clot. The first target location is selected by the clinician, which is typically the furthest end of the long-axis of the hematoma from the robot. The robot is inserted to that location, and the dynamic scan sequence at ~1 MRI frame/s is used to validate that the robot moves along the projected path and that the target location is reached. An iterative evacuation loop would then be performed. In this loop, the clinician adjusts the position of the robot based on the imaging feedback provided by the MRI until an acceptable residual hematoma (< 15 mL) remains (~5 minutes) [18]. The robot is then removed from the patient once the clinician is satisfied with the evacuation outcome.

## V. PLASTIC TUBE SELECTION AND MODELING

### A. Forward Kinematics

Although the well-known constant curvature assumption provides an elegant solution for kinematic modeling, the use of plastic tubes reduces modeling accuracy significantly due to variability in curvature across the length of the tube (tip position RMSE max of 5.86 mm in our robot). This section presents a new single image characterization method for analyzing the variable curvature of concentric tubes to improve the resulting error (tip position RMSE max of 2.05 mm in our robot). Note that we assume the outer tube possesses zero curvature and a sufficiently high elastic and torsional modulus, removing it from consideration of the model. This is an accurate assumption as the outer tube (6394A14, McMaster-Carr, GA, USA), made of fiberglass, possesses an elastic modulus of 74 GPa and a shear modulus of 30 GPa, whereas the inner nylon tube possesses an elastic modulus of 4 GPa and a shear modulus of 2.7 GPa.

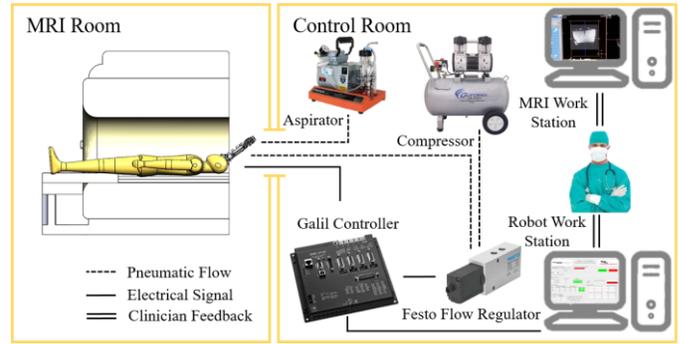

**Fig. 7.** The robot's procedural setup in the MRI scanner and control room.

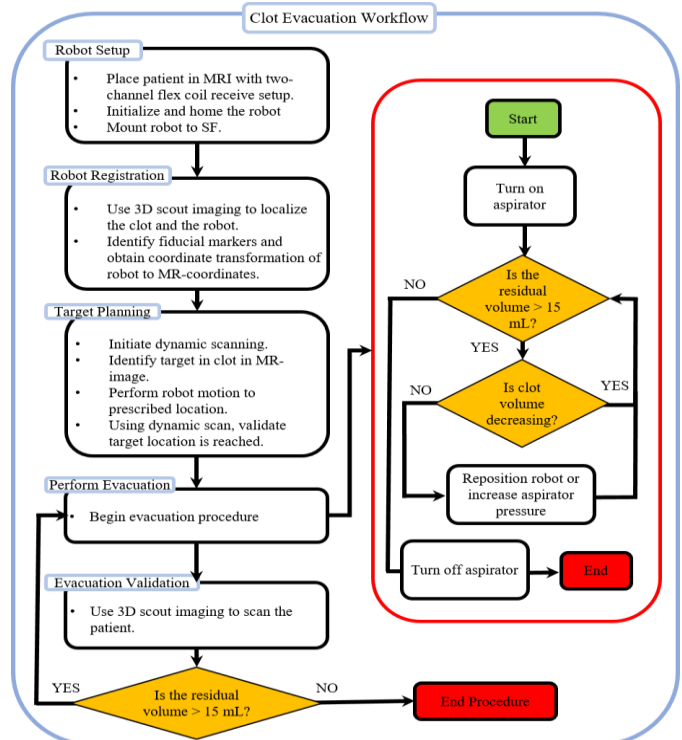

**Fig. 8.** The proposed in-scanner clinical workflow. Note that this procedure is after the SF has been mounted to the patient's skull. Thus, the robot's initial insertion vector is defined. The iterative loop would provide real-time feedback of the clot's evacuation at ~1 Hz.

The model begins by first representing the inner tube's shape as a function $f(x)$, as shown in Fig. 9. The origin of the function exists at the distal end of the outer tube and the maximum value of $x$ occurs at the distal end of the inner tube at its maximum displacement from the outer tube. We assume the inner tube's first- and second-order derivatives are continuous, which is true since the inner tube continuously bends along the body. Thus, the curvature and arc length at any point along the function can be defined by:

$$\kappa(x) = -\frac{f''(x)}{(1+f'(x)^2)^{\frac{3}{2}}} \tag{3}$$

$$ds(x_1, x_2) = \int_{x_1}^{x_2} \sqrt{1+f'(x)^2}\,dx \tag{4}$$

where $f'(x)$ and $f''(x)$ are the first and second derivative of the function, $x$ is the independent variable of the function, $\kappa(x)$ is the curvature of the function at some point $x$, and $x_1$ and $x_2$ are the lower and upper bounds of integration used to determine the arc length for a discrete segment $ds$ of the function.

In order to obtain the tube shape with respect to the insertion depth $s$, the lower and upper limit of $x$ must be determined. The upper limit of $x$ occurs at the distal tip of the tube, which is $x_{max}$, as shown in Fig. 9. However, the lower bound of integration $x_{min}$ decreases with increasing $s$. In some instances, it is possible to solve for $x_{min}$ in terms of $s$ by evaluating the integral defined by (4), where $x_1$ would be the unknown, $ds$ would be equal to $s$, and $x_2$ would be equal to $x_{max}$. However, this is often an intractable approach. Instead, we solve for $x_{min}$ numerically by driving the left-hand side of (5) to zero using the fzero function in MATLAB, where $x_{max}$ is known and $x_{min}$ is the variable.

$$\int_{x_{min}}^{x_{max}} \sqrt{1+f'(x)^2}\,dx - s = 0 \tag{5}$$

Using (3) and (4), the change in shape $dT$ of a discrete element of $f(x)$ can be represented by the transformation matrix:

$$dT(\kappa(x_1), ds(x_1, x_2)) =$$
$$\begin{bmatrix} 1 & 0 & 0 & 0 \\ 0 & \cos(\kappa\,ds) & -\sin(\kappa\,ds) & \frac{\cos(\kappa\,ds)-1}{\kappa} \\ 0 & \sin(\kappa\,ds) & \cos(\kappa\,ds) & \frac{\sin(\kappa\,ds)}{\kappa} \\ 0 & 0 & 0 & 1 \end{bmatrix} \tag{6}$$

Assuming that for an infinitesimal segment of the tube the curvature is constant, the tube can be discretized into $n$ points (100 in this work) and the total shape of the tube at any point $x$, given an insertion depth $s$, can be expressed as:

$$T_1^n(s) = \prod_{j=1}^{n} dT\left(\kappa(x_{j-1}), ds(x_{j-1}, x_j)\right) \tag{7}$$

where $x_j$ is defined by:

$$x_j = x_{min} + \frac{j}{n}(x_{max} - x_{min}) \tag{8}$$

It should be noted that this evaluation can occur for any planar tube shape. Given (7), the CTR's full configuration can be expressed as:

$$T_0^n = T_0^1 T_1^n(s) \tag{9}$$

where $T_0^1$ is defined as:

$$T_0^1 = \begin{bmatrix} \cos(\theta) & -\sin(\theta) & 0 & 0 \\ \sin(\theta) & \cos(\theta) & 0 & 0 \\ 0 & 0 & 1 & d \\ 0 & 0 & 0 & 1 \end{bmatrix} \tag{10}$$

where $d$ is the translational displacement of the outer tube from the SF and $\theta$ is the rotational displacement of the inner tube around its longitudinal axis. The resulting total configuration given the joint space parameters $d$, $s$, and $\theta$ and their corresponding coordinate systems can be seen in Fig. 10.

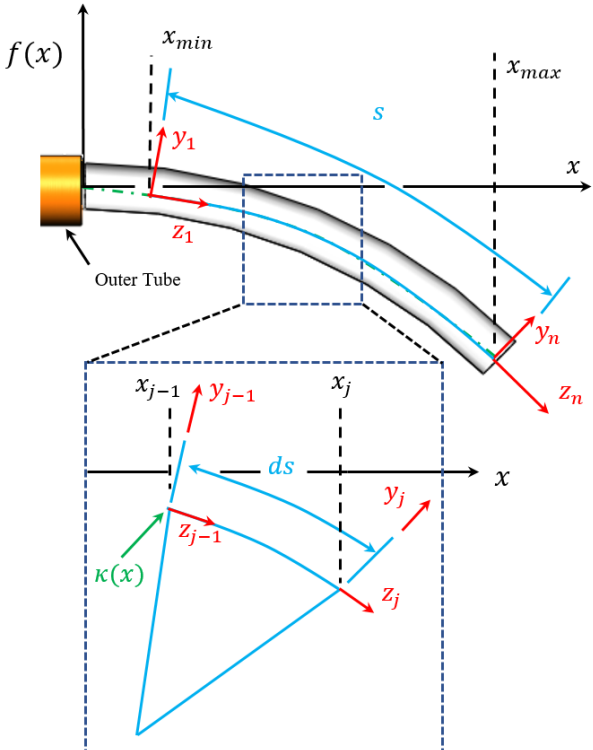

**Fig. 9.** The modeling strategy can be seen superimposed upon a pre-curved tube of arbitrary shape defined by the function $f(x)$.

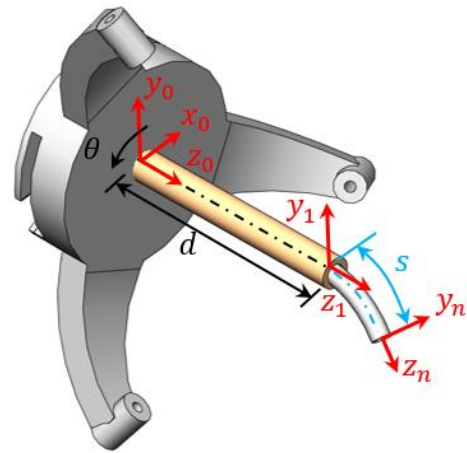

**Fig. 10.** The inner and outer tubes can be seen extending from the SF. Their corresponding coordinate systems (red) and their joint space parameters (black and blue) can be seen superimposed on the inner and outer tubes.

## B. Torsional Modeling

Our inner plastic tube has a low shear modulus (< 3 GPa). Consequently, when the inner tube is inside the outer tube, the force applied by the outer tube that straightens the inner tube produces a resistive torque when the inner tube is rotated, resulting in a tip position RMSE as high as 1 mm from torsional displacement alone. Thus, we develop a model that considers and compensates for the torsional deflection of the inner tube caused by the outer tube. Similar to our derivation for the forward kinematics, we elect to use a function $f(x)$ to describe the inner tube's shape.

We begin by considering a discrete section of the tube bent with curvature $\kappa$ as a cantilever beam and calculate the force necessary to straighten the discrete section, as shown in Fig. 11. We define the strain energy in this discrete section of the tube using Castigliano's first theorem, which can be written as [62]:

$$\frac{\partial U}{\partial \kappa} = FL \tag{11}$$

where $U$ is strain energy stored in the system, $\kappa$ is the curvature of the tube, $F$ is the force applied to change the curvature of the discrete segment, and $L$ is the moment arm of the applied force. To solve for the strain energy associated with a discrete section of the tube, it is first recognized that the strain $\varepsilon$ in an axial fiber of any cross-section based on curvature $\kappa$ can be defined by [34]:

$$\varepsilon(y, \kappa) = \frac{\kappa(y - \bar{y})}{1 + \bar{y}} \tag{12}$$

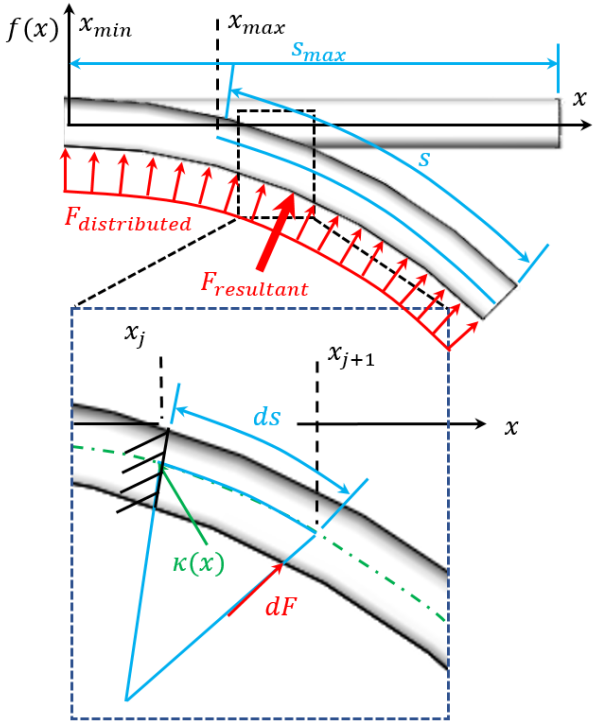

**Fig. 11.** The modeling strategy for analyzing torsion can be seen. This strategy relies on assuming that the distributed load (red) applied across the inner tube's body can be broken up into discrete elements. Note that for the small segment $ds$, the force is applied at the distal end while the proximal end is assumed to be held fixed as ground for modeling purposes, isolating the discrete segment.

where $y$ is the distance from a coordinate system defined in the cross-section of the rod and $\bar{y}$ is the distance of the neutral bending plane from the same coordinate system. Using (12), the strain energy density within an axial fiber of the discrete section is equal to the area under the stress-strain curve of that fiber, written as:

$$W(\varepsilon(y, \kappa(x))) = \int_0^{\varepsilon} \sigma(e)de \tag{13}$$

where $W$ is the strain energy density as a function of strain and curvature, as defined by (12) and (3), and the stress $\sigma$. As a result, the strain energy induced by straightening a segment $ds$ from its precurved configuration can be defined as:

$$dU(x_1, x_2) = ds(x_1, x_2) \int_{r_{ID,i}}^{r_{OD,i}} \int_0^{2\pi} W(\varepsilon(y, \kappa(x_1)))d\varphi dr \tag{14}$$

where $r_{OD,i}$ and $r_{ID,i}$ are the outer and inner radius of the inner tube and $ds$ is the length of the discretized segment, as shown in the detailed view in Fig. 11. Using (14), we can rewrite (11) in its discretized form to find the force applied to that segment as:

$$dF(x_1, x_2) = \frac{1}{ds(x_1, x_2)} \frac{\partial dU(x_1, x_2)}{\partial \kappa(x_1)} \tag{15}$$

where $ds$ is the moment arm. Using (15), the sum of all the discrete forces, i.e. the distributed load applied to the deflectable region of the inner tube when inside the outer tube, can be written as a function of insertion depth $s$ as:

$$F(s) = \sum_{j=1}^{n} dF(x_j, x_{j+1}) \tag{16}$$

where $n$ is the number of discrete segments (100 herein) and $x_j$ is defined as:

$$x_j = \frac{j}{n}(x_{max}) \tag{17}$$

where $x_{max}$ in this section increases with insertion depth, as shown in Fig. 11. $x_{max}$ can be numerically solved for by driving the left-hand side of the following to zero:

$$\int_0^{x_{max}} \sqrt{1 + f'(x)^2}dx - s_{max} - s = 0 \tag{18}$$

It should be noted that (16) is the resultant force of the distributed load applied to the inner tube's surface by the outer tube. By considering the resultant force to occur at a single point, the resistive torque can be solved for as:

$$T(s) = \mu F(s)r_{OD,i} \tag{19}$$

where $T(s)$ is the torque applied to the tube and $\mu$ is the coefficient of static friction between the inner and outer tube, 0.17 in this case. However, before torsional deflection can be evaluated, the location of the resultant force must be determined, which can be expressed as [62]:

$$L_{resultant}(s) = \frac{\sum_{j=1}^{n} dF(x_j, x_{j+1}) ds(0, x_j)}{F(s)} \quad (20)$$

where $L_{resultant}(s)$ is the location of the resultant force. Note that $L_{resultant}(s) = 0$ occurs at the proximal end of the deflectable region of the inner tube. Using (19) and (20), the angular deflection induced by an applied torque can be defined as:

$$\varphi(s) = \frac{T(s)(L_i - L_{resultant}(s) - s)}{JG} \quad (21)$$

where $\varphi(s)$ is the angular deflection in radians between the base of the inner tube (at the polygon spline) and the location of the resultant force, $L_i$ is the total length of the inner tube, $J$ is the polar second moment of inertia, and $G$ is the shear modulus of the inner tube. Using $\varphi(s)$, the torsional deflection of the inner tube can be compensated, improving targeting accuracy.

### C. Inverse Kinematics

For inverse kinematics, we include torsional deflection compensation in the model. It should be noted that the amount of torsional deflection compensation is dependent on the position before and after a desired move. Thus, the following logic is used to compensate:

$$\theta(s) =$$
$$\begin{cases} \theta_{nom} + sign(\theta_{nom} - \theta_{curr})(\varphi(s_{nom})) & s_{curr} < s_{nom} \\ \theta_{nom} + sign(\theta_{nom} - \theta_{curr})(\varphi(s_{curr})) & s_{curr} \geq s_{nom} \end{cases} \quad (22)$$

where $\theta$ and $s_{nom}$ are the nominal rotation angle and insertion displacement of the inner tube for the next position calculated using inverse kinematics, $\theta_{curr}$ and $s_{curr}$ are the robot's current nominal rotation angle and insertion displacement, and $sign(\theta_{nom} - \theta_{curr})$ provides the sign of the difference between the two angles, identifying the direction of rotation of the next move.

In this robot a move is performed by first rotating the inner tube at its current position given the above calculations and then translating to its prescribed depth. Using this logic, the amount of torsional deflection compensation $\varphi(s)$ is dependent on the minimum value of $\varphi(s)$ between $s_{curr}$ and $s_{nom}$. For example, if the inner tube is extending 20 mm from the outer tube and in the new position it will extend only 10 mm, we will use the lower part of the logic in (22), i.e., we will rotate the inner tube at the current position and compensate for torsional deflection based on the current position. This is because retracting the tube does not apply a torque to the tube, so the compensation is only dependent on the torsional deflection prior to retraction. Conversely, if the inner tube is extending 10 mm from the outer tube and in the new position it will extend 20 mm, we will use the upper part of the logic in (22), i.e., we will rotate the inner tube at the current position, but compensate for torsional deflection based on the new position. This is because at the new position the additional section of the inner tube extending from the outer tube no longer experiences torsional deflection defined by (21), meaning compensation for this section of the tube is unnecessary. Note that this model assumes that the forces applied by the clot are negligible and the outer tube acts as a rigid delivery system for the inner tube, negating forces potentially applied by the phantom/brain.

### D. CTR Tube Selection

The clinical requirements of this project, specifically the inner tube's evacuation diameter, limit the scope of material selection to materials with low elastic moduli ($< 8$ GPa). This is justified by considering the tube used in our prior work in [63]. The tube used was nitinol with the design parameters shown in Table 1 row 1. Using our force analysis model described in Sec V-C, these design parameters result in a bending stiffness of $82.84 \frac{N}{rad}$ and a total force of 111.84 N (applied at the tube's tip) needed to bend the tube into a straight configuration. However, if this tube was to be increased to the diametral dimensions of the clinical requirements matching our nylon tube (Table 1 row 2), the resulting bending stiffness would be $1568.3 \frac{N}{rad}$, requiring a total force of 2117.2 N to bend the tube into the straight configuration. This analysis clearly illustrates the limitations of metallic tubes for this clinical application. To ameliorate the large increase in bending stiffness associated with the diameter sizing, highly elastic nylon tubing was used for the inner tube in lieu of the typical nitinol tube used in CTRs. Using nylon, a tube with the same design parameters as [63], but with our diametral requirements met, would result in a bending stiffness of $79.58 \frac{N}{rad}$ and a total force of 107.44 N to bend the tube into the straight configuration.

**Table 1.** Comparison of Bending Stiffness to Tube Properties

| Tube Material | Modulus of Elasticity [GPa] | Outer Diameter [mm] | Inner Diameter [mm] | Radius of Curvature [mm] | Length of Curvature [mm] | Stiffness [N/rad] | $F_{bend}$ [N] |
|---|---|---|---|---|---|---|---|
| Nitinol[1] | 74 | 2.2 | 1.5 | 20 | 27 | 82.84 | 111.84 |
| Nitinol[2] | 74 | 6 | 4 | 20 | 27 | 1568.3 | 2117.2 |
| Nylon[3] | 4 | 6 | 4 | 20 | 20 | 79.58 | 107.44 |
| Nylon[4] | 4 | 6 | 4 | $\frac{1}{\kappa(x)}$ | 29 | 79.58 | 66.46 |

1. The nitinol tube used in [63].
2. A nitinol tube with the diameter requirements of our tube, but the same RoC and LoC of (1).
3. The materials properties of our tube with the same RoC and LoC of (1).
4. Our tube's design parameters.

Plastic materials provide the unique advantage of a lower heat-setting temperature ($< 200$ °C) as compared to their metallic counter-parts ($> 500$ °C). This allows this ability to fabricate patient specific curved tubes with any desired curvature based on the pre-operative scan in under 5 minutes. We fabricate our plastic tubes using a heat-gun from a soldering station and a PLA printed fixture. The fixture has pre-defined RoCs ranging from 20-50 mm in 5 mm increments. To shape the tube, the heat-gun is brought to a temperature of 350 °C. The desired LoC of the tube is heated using the heat gun for 30 seconds and is then placed in the fixture at the desired RoC. This tube is allowed to sit for 3 minutes in the fixture and is then removed and cooled under running water. This rapid manufacturing process can allow a clinician to make modifications to the inner tube at any point in the procedure. However, it should be noted that plastic materials tend to undergo plastic deformation, a state where the stress induces a strain that is sufficient to permanently deform a material. CTRs rely on the inherent super-elasticity of pre-curved tubes to expand the workspace compared to the conventional straight devices. Herein, we determine the maximum initial curvature our inner tube can have given the material's elastic strain limit, which is an extension of prior work [64].

Commonly, residual stresses and strains are removed from pre-curved tubes using heat treatment. Therefore, the strain of a pre-curved tube can be written using (12) in its unloaded condition as:

$$\varepsilon_o(y, \kappa_o) = \frac{\kappa_0(y - \bar{y})}{1 + \bar{y}\kappa_0} = 0 \qquad (23)$$

where $\varepsilon_o$ is the initial strain, which is zero after heat treatment, and $\kappa_0$ is the initial curvature after heat treatment. In this work, the tube used is axis-symmetric. Therefore, it is convenient for analysis to place the origin of the coordinate system at the center of the circular cross-section, equating $\bar{y}$ to 0. Therefore, the strain in a pre-curved tube undergoing a change in curvature can be expressed as:

$$\varepsilon(y, \Delta\kappa) = \Delta\kappa(y) \qquad (24)$$

where:

$$\Delta\kappa = \kappa_0 - \kappa \qquad (25)$$

Note that $\kappa$ can be positive or negative. For convention, positive curvature curves the tube into the negative y-direction, as shown in View B-B of Fig. 12. The maximum allowable change in curvature occurs when strain is at its maximum limit within the elastic range from its initial configuration. For a cross-section in pure bending, the maximum strain always occurs at the fiber furthest from the neutral bending plane, i.e. the outer radius for a circular cross-section. Therefore, (24) can be rewritten as:

$$\varepsilon_{limit}(r_{OD}, \Delta\kappa) = (\kappa_0 - \kappa_{limit})(r_{OD}) \qquad (26)$$

where $\kappa_{limit}$ is the maximum change in curvature given an elastic strain limit $\varepsilon_{limit}$ and $r_{OD}$ is the outer radius of the tube

in question. Rearranging (26) and solving for the initial curvature $\kappa_0$, we obtain:

$$\kappa_0 = \pm \frac{\varepsilon_{limit}}{r_{OD}} + \kappa_{limit} \qquad (27)$$

Note that $\frac{\varepsilon_{limit}}{r_{OD}}$ can be positive or negative, indicating that $\kappa_0$ has two solutions. Herein, we focus on the solution that results in $\kappa_0$ in the negative y-direction, i.e. $+\frac{\varepsilon_{limit}}{r_{OD}}$. $\kappa_{limit}$ could be assumed to be zero. In this application, this would occur when the inner tube is fully retracted into the outer tube with the inner diameter of the outer tube equal to the outer diameter of the inner tube. However, this is not always the case. In some instances, the outer tube may possess an inner diameter larger than the outer diameter of the inner tube. If this condition exists, an expression for $\kappa_{limit}$ can be obtained by first including two assumptions:

1. The inner tube's deformation abides by the constant curvature assumption, as depicted in Fig. 12, View A-A. Note that this assumption is only used for determining the maximum curvature.
2. The inner tube is in contact with the outer tube at the distal end of the inner tube and the proximal end of the curved region of the inner tube.

The first assumption permits implementation of the constant curvature transformation matrix defined below [31, 34]:

$$T(\kappa, s_{max}) = \begin{bmatrix} 1 & 0 & 0 & 0 \\ 0 & \cos(\kappa s_{max}) & -\sin(\kappa s_{max}) & \frac{\cos(\kappa s_{max}) - 1}{\kappa} \\ 0 & \sin(\kappa s_{max}) & \cos(\kappa s_{max}) & \frac{\sin(\kappa s_{max})}{\kappa} \\ 0 & 0 & 0 & 1 \end{bmatrix} \qquad (28)$$

where $s_{max}$ is the length of the deflectable region of the tube. The second assumption defines the transformation matrix from the centerline of the tube at its distal end to the inner radius of the outer tube, represented by (29), and the transformation matrix to the final target position, represented by (30).

$$T_{y,r} = \begin{bmatrix} 0 & 0 & 0 & 0 \\ 0 & 0 & 0 & -r_{OD,i} \\ 0 & 0 & 0 & 0 \\ 0 & 0 & 0 & 1 \end{bmatrix} \qquad (29)$$

$$T_0^2 = \begin{bmatrix} n_x & s_x & a_x & d_x \\ n_y & s_y & a_y & d_y \\ n_d & s_d & a_d & d_z \\ 0 & 0 & 0 & 1 \end{bmatrix} \qquad (30)$$

where $r_{OD,i}$ is the outer radius of the inner tube. Referring to Fig. 12, View A-A, the total displacement in the y-direction can be defined by:

$$d_y = D_{ID,o} - r_{OD,i} \qquad (31)$$

where $D_{ID,o}$ is the inner diameter of the outer tube and $d_y$ is the target position in the y-coordinate. Using inverse kinematics, $\kappa_{limit}$ can be solved numerically by driving the left-hand side of (32) to 0.

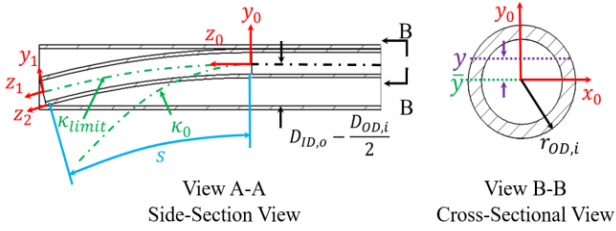

**Fig. 12.** View A-A depicts a section view of a concentric tube pair with the inner tube withdrawn into the outer tube. In this configuration, the inner tube has a significantly smaller outer diameter than the outer tube's inner diameter to aid in visualization. View B-B depicts the cross-section of the inner tube and the coordinate system defined for evaluating strain based on curvature.

$$\frac{\cos(\kappa_{limit}s)-1}{\kappa_{limit}} - r_{OD,i}\cos(\kappa_{limit}s) = 0 \tag{32}$$

In our application, the inner diameter of the outer tube and the outer diameter of the inner tube are equivalent, driving $\kappa_{limit}$ to zero. The inner tube's material was highly elastic nylon (50405K35, McMaster-Carr, GA, USA), providing a $\varepsilon_{limit} = 10\%$. Based on the previous analysis, the anticipated minimum RoC is 30 mm. However, it should be noted that spring-back, the tendency of a bent material to revert to its original form after shaping, is a prevalent issue during the heat-setting process of any material with high strain limits [65]. Thus, when forming our tube, we apply an initial RoC of 20 mm using the manufacturing method discussed at the beginning of this section (Fig. 13A-B). We then place the inner tube inside the outer tube for 1 hour after cooling (Fig. 13C). The 1 hour straightening of the tube ensures the final shape is dependent upon the relationship described by the equations above. After the tube was removed from the outer tube, an image was taken of the inner tube and the resulting average RoC was 32.4 mm (Fig.13D). The fixture and inner tube prototype can be seen in Fig. 13E. To validate whether the inner tube can return to the pre-shaped configuration during operation, the inner tube prototype was cycled in and out of the outer tube for 1000 cycles with a translational lead-screw mechanism. An image was taken after cycling and the resulting average RoC of the inner tube was calculated as 33.4 mm. This validated the equations discussed above and ensures that plastic tubes are a viable option for concentric tube robots.

## VI. Experimental Procedure and Results

### A. MR-conditional Pneumatic Motor Control

The voltage values for starting a stationary motor $V_{start}$ and avoiding the dead-band region $V_{d-b}$ were measured experimentally for the motor controlling each axis. Experiments were performed with the robot in its assembled state, ensuring that system dynamics (such as viscous damping of the planetary gearboxes) were included in the characterization of the start and dead-band voltages. The motors were driven using a compressor (8010, California Air, CA., USA.) with a max pressure of 90 psi. While measuring the start and dead-band voltages for a given axis, the other two axes remained stationary. For each axis, the starting voltage was measured by incrementing the offset voltage applied by the Galil motion controller by 0.1 V, starting at 0 V. The voltage value that resulted in the motor for that axis achieving a non-zero rotational velocity was recorded as $V_{start}$. In a similar fashion, $V_{d-b}$ was measured by incrementing the offset voltage applied by the Galil motion controller by -0.1 V, starting at $V_{start}$. The voltage value prior to the zero rotational velocity was recorded as $V_{d-b}$ (i.e. if the motor stops rotating at 2.9 V, $V_{d-b} = 3.0$ V). This was done for both clockwise and counter-clockwise rotation of the motor, i.e. voltages from 0 to10 V and 0 to -10 V. For the motors controlling the translational axes of the inner and outer tube, the resulting start voltage was $\pm 6.4$ V and the resulting dead-band voltage was $\pm 3.0$ V. For the motor controlling the rotational axis of the inner tube, the start voltage was $\pm 9.0$ V and the dead-band voltage was $\pm 3.5$ V.

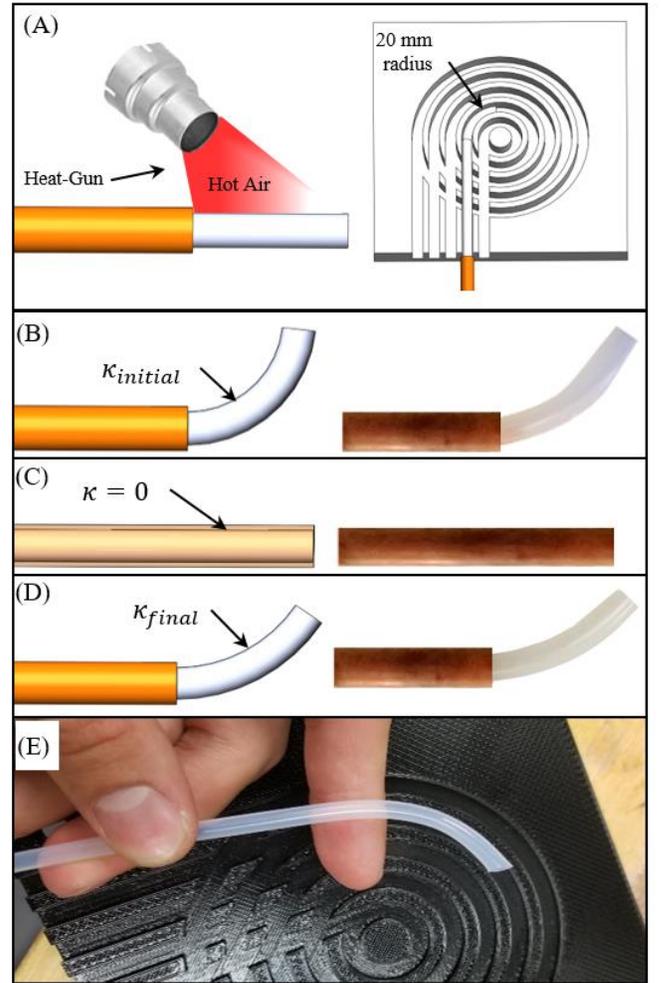

**Fig. 13.** (A) The inner tube's curved portion is produced by heating the distal end of the tube using a heat-gun and placing it in the fixture. (B) A radius of curvature less than the minimum radius of curvature is first produced and the tube is allowed to cool. (C) The inner tube is retracted into the outer tube and remains so for one hour, inducing plastic deformation. (D) The plastic deformation results in the final radius equal to the minimum radius of curvature described by Section V-D. (E) An example of the printed fixture and curved tube.

Using the voltage characterization values in the control law, the positional accuracy of the motors were tested with the robot in the assembled state. $err_{thresh}$ was set to 90° (1000 encoder counts) based on qualitative observations that this improved response time without adversely affecting overshoot. This resulted in a $K_P$ in region II of 9.8 and 11.5 at $err = 90°$ for the translational and rotational axes, respectively. Thus, $K_{P,const}$ was set to 9.8 for the motors controlling translation and 11.5 for the motor controlling rotation, providing a smooth transition between the constant and variable regions of the control law. The integral and derivative gains were 0 and 4000, respectively, for all experiments in this work. $\delta$ was set to 1°, preventing the valves from continuously leaking air pressure when sufficiently close to the setpoint, while maintaining a sufficient accuracy for our work. These parameters resulted in a motor accuracy of 0.317°±0.3°, which is an improvement compared to our prior work (1° error) [60]. An example of the motor used for the rotational degree of freedom (gear reduction of 400:1) transitioning between the constant PID and variable portion of the control law region can be seen in Fig. 14A. A set point of 144° was prescribed at $t = 0.846$ s. The motor began to respond at 1.56 s and obtained a steady state error of zero at $t = 8$ s. Note the corresponding voltage signal provided by the control law in Fig. 14B.

### B. Kinematic Model Validation

The kinematic model was validated in free-space using a camera (5WH00002, Microsoft LifeCam Web Camera) placed perpendicular to the longitudinal axis of the robot. The camera was calibrated using the MATLAB Camera Calibration Toolbox [66]. The inner tube's shape was first characterized with the inner tube at an insertion depth of 29 mm (maximum insertion depth difference) with respect to the outer tube. A single picture of the tube was taken in this configuration and a $4^{th}$ order polynomial $f(x)$ was fitted to the center-line of the tube shape, as discussed in Section V-A. The inner tube was then configured with an initial insertion depth of 0 mm with respect to the outer tube (fully retracted), and inserted in increments of 1 mm up to 29 mm. At each increment, an image of the tube shape was recorded and points were selected along the centerline of the tube to obtain the experimental tube shape. The model was then used to obtain the tube shape at each increment. Examples of the tube shape using the model can be seen in Fig. 15A.

The results of the model and the constant curvature assumption (shape seen in Fig. 15A) at each insertion depth were both compared to the experimental images. The curvature for the constant curvature plot was assumed to be equal to the average curvature of the tube $\kappa_{const} = \frac{1}{33.4} mm^{-1}$ and the maximum insertion depth was considered to be the same $s_{max}$ = 29 mm. Table 2 presents the results of the two models to the experimental results. The metrics of comparison were the average RMSE of the tip position of the tube and the average RMSE of the tube's shape ± one standard deviation. The RMSE of the tube's shape is defined as the average of the minimum RMSE at 1 mm increments along the tube's body. For example, at $s = 10$ mm, the minimum RMSE between the model in question and the experiment would be calculated 10 times, one for each 1 mm increment, and then averaged. Note that the average RMSE and standard deviation do not reflect the maximum errors seen, which were 2.04 mm for our model and 5.86 mm for the constant curvature model, highlighting our model's efficacy (Fig. 15B).

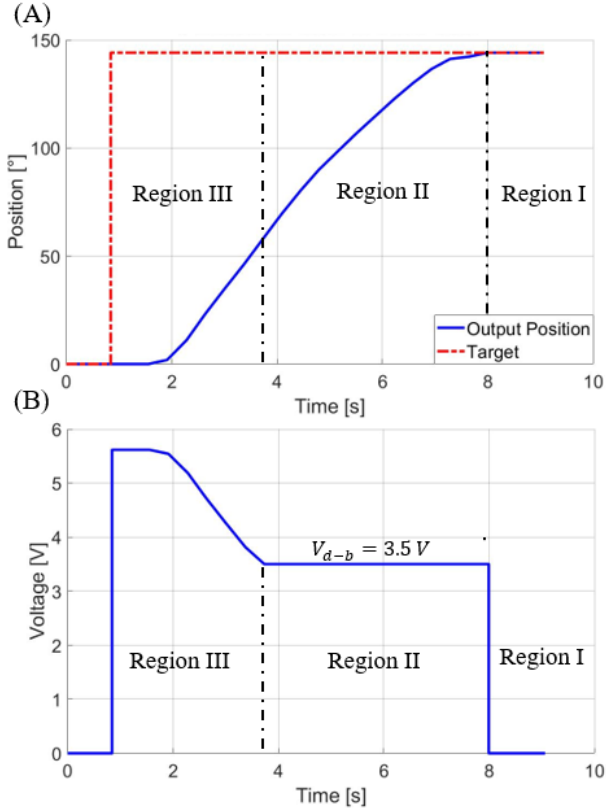

**Fig. 14.** (A) The motor's response to a setpoint of 144° can be seen with a resulting steady-state error of 0°. (B) The voltage signal supplied to the motor in (A) based on the control law can be seen with a dead-band voltage of 3.5 V. The constant voltage in region III is due to system delay, whereas the constant voltage in region II is due to the control-law varying gain to maintain a voltage level above the dead-band region without steady-state oscillatory behavior.

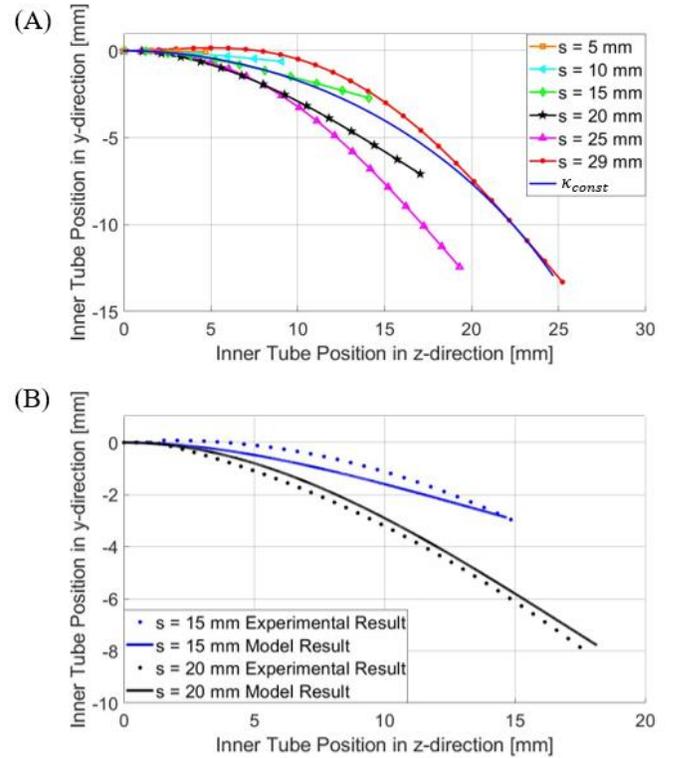

**Fig. 15.** (A) depicts the tube shape for different insertion depths calculated using the model discussed in Section V-B. Note that the plot of constant curvature can also be seen with $\kappa_{const} = \frac{1}{33.4} mm^{-1}$. (B) shows a comparison between the experimental results and the model of the tube shape at an insertion depth of 15 and 20 mm.

**Table 2.** In-Plane Comparison of the Constant Curvature and Characterization Modeling Approaches

| Insertion Depth [mm] | Tip RMSE [mm] (Constant Curvature) | Robot Shape RMSE [mm] (Constant Curvature) | Tip RMSE [mm] (Characterized Method) | Robot Shape RMSE [mm] (Characterized Method) |
|---|---|---|---|---|
| $s = 5$ | 1.03 | 0.30±0.11 (Max: 0.61) | .06 | 0.14±0.05 (Max: 0.28) |
| $s = 10$ | 1.27 | 0.43±0.23 (Max: 1.12) | 0.80 | 0.30±0.16 (Max: 0.84) |
| $s = 15$ | 1.22 | 0.34±0.22 (Max: 1.09) | 0.16 | 0.36±0.20 (Max: 0.62) |
| $s = 20$ | 2.00 | 1.01 ±0.69 (Max: 2.12) | 0.65 | 0.31±0.17 (Max: 0.66) |
| $s = 25$ | 4.47 | 2.08±1.80 (Max: 5.86) | 1.32 | 0.67±0.25 (Max: 2.04) |
| $s = 29$ | 1.61 | 0.73±0.42 (Max: 1.56) | 0.46 | 0.25±0.14 (Max: 0.80) |
| Average | 1.93 | 0.82 | 0.58 | 0.34 |

### C. Free-Space Modeling Validation

A set of systematic experiments were performed in free-space to validate both the accuracy of each degree of freedom individually, as well as the accuracy of the system as a whole. In this study, the inner tube's tip position was recorded using an Electro-Magnetic (EM) field generator (Aurora, NDI Medical, Ontario, Canada) and its 6-DoF probe sensor, as shown in Fig. 16. The robot was rigidly mounted to the table and its coordinate frame was registered to the tracker coordinate frame through the point-based registration method [61]. The translational degrees of freedom was validated by translating both tubes to a prescribed insertion depth, recording the resulting position, then retracting both tubes to a different prescribed insertion depth and recording the resulting position again. The absolute insertion depth ranged between 30 and 60 mm and the absolute retraction depth ranged between 5 and 10 mm. This procedure was repeated 20 times, resulting in an RMSE of 0.24±0.21 mm.

The rotational degree of freedom was validated by extending the inner tube to its maximum insertion depth (29 mm) with respect to the outer tube, eliminating the application of a resistive torque, and rotating the tube clockwise and counter-clockwise in increments of 36 degrees. This test was repeated 20 times, resulting in an average tip RMSE of 1.09±0.67 mm, indicating an angular error of 3.1±2.36°.

After confirmation that the system hardware used for each degree of freedom provided sufficient accuracy, the whole system was validated using the inverse kinematics model in Section V-C, accounting for both torsional compensation and the variable curvature of the tube. Forty-eight target points were selected in the robot's coordinate frame. These targets were selected randomly to provide unbiased random targets. The x and y target location ranged between 8 and -8 mm and the z target location varied between 20 and 70 mm. The robot was able to maintain a targeting RMSE of 1.39±0.54 mm (Max: 2.46 mm), which is an improvement over the case when torsion is not considered (RMSE 1.66±0.91 mm (Max: 4.32 mm)) and well within our desired RMSE of 3.2 mm.

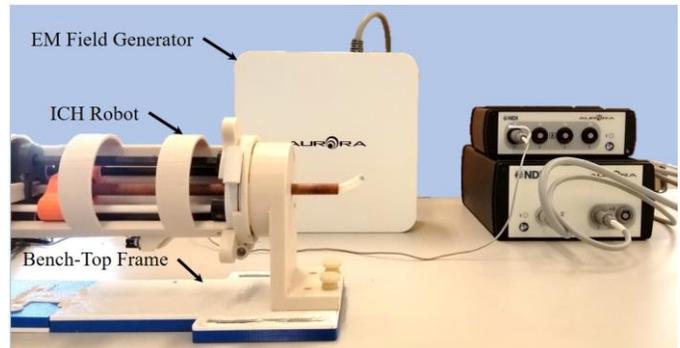

**Fig. 16.** Bench-top setup used to validate the robot's forward and inverse kinematics.

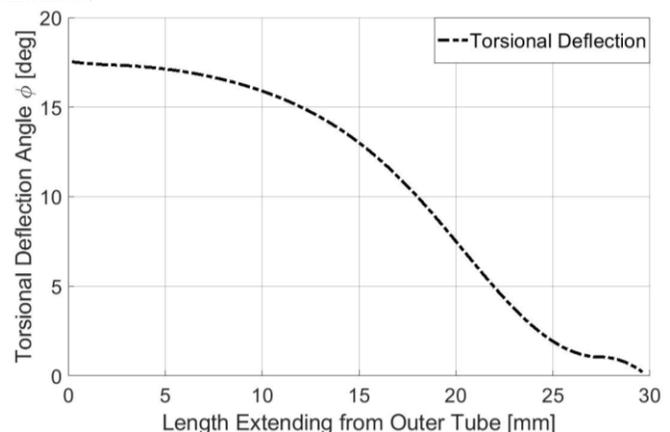

**Fig. 17.** Results of the anticipated torsional deflection given insertion depth and inner tube rotation as discussed in Section V-C based on the 4th-degree polynomial defined by the needle shape characterized in Section V-A.

Note that the torsional compensation model is highly non-linear, as depicted in Fig. 17. However, the torsional model does behave as expected based on the tube shape. For example, the tube shape is relatively straight at the distal and proximal end of the deflectable region, as shown by the insertion depth of 29 mm in Fig. 15A. Thus, the change in resistive torque and torsional deflection should be small at the beginning and end of the insertion depth, which is clearly captured by Fig. 17. Note that although the torsional model was not directly validated, it was indirectly validated by the accuracy of the robot's inverse kinematics.

### D. MRI Tube Artifact and Robot Conditionality

An imaging experiment was performed to provide a comparison of the signal void artifact produced by the selected plastic (6 mm OD) and fiberglass (7.93 mm OD) tubes to the traditional nitinol tube (3 mm OD). A fixture was made to hold the tubes inside a water phantom with their longitudinal axis parallel to the transverse plane and perpendicular to the magnetic field. 3D spoiled gradient-echo imaging (FOV: $240 \times 240 \times 240$ mm$^3$, $0.9 \times 0.9$ mm$^2$ resolution, TR/TE = 5/2 ms) was performed to visualize the tube artifacts produced by the difference in magnetic susceptibilities of the tube material and surrounding water, as shown in Fig. 18A. To provide a metric of comparison, a slice in the transverse plane was selected where the artifact size was maximum. The size of the artifact was then compared to the original tube's size. The artifact of the outer fiberglass tube, inner plastic tube, and sample nitinol tube each had a percent increase of 2.14%, 6.07% and 440%, respectively, compared to their original diameter. This clearly indicates the need for non-metallic tubes for ICH evacuation with MR-guidance.

The robot's MR-conditionality was validated using 3D $T_1$ weighted imaging (Fig. 18B). In this experiment, a bottle water-phantom was located at the isocenter of the MRI scanner and a region of interest (ROI) within the bottle was compared under three scenarios: (i) no robot in the MRI scanner, (ii) robot in the MRI-scanner placed near the bottle and turned off, and (iii) robot in the MRI-scanner and operating. The ROI was a circular region with a diameter of 100 pixels selected in the coronal plane. An artifact can be defined as 30%-pixel intensity change within the ROI image [67]. In our case, there was less than 1% change in intensity for both the robot off and robot operating conditions. Note that different ROI planes and pixel diameters were selected and the difference in the change in intensity remained below 1%.

### E. MRI Phantom Accuracy Characterization

The robot's tip accuracy was validated in a 70 cm wide bore, Philips 3 Tesla Ingenia Elition MRI System using a 10% by weight Knox™ (Kraft Foods Global, Inc., USA) gelatin phantom, as shown in Fig. 19A. During the experiment, the phantom was placed in front of the robot with a pair of flexible imaging coils on either side of the phantom (Philips dStream Flex-S 2 channel receive coils). The robot was first localized using MR visible fiducials (PinPoint® #128, Beekley, USA) placed on the front of the bench-top frame using scout imaging. Following localization, ten target locations were selected at different depths within the phantom around the rotational axis of the inner tube in the MR coordinate system. These targets were then converted to the robot's coordinate frame and inverse kinematics were used to determine the joint space parameters needed to reach the target. An iterative ~1 Hz single slice scan was performed in the plane of the inner tube's curvature based on the angle of rotation determined from the inverse kinematics using the dynamic imaging protocol discussed in Section IV-E. The robot was moved to that position while continuous 2D scanning was performed for visualization and accuracy validation, as shown in the time sequence in Fig. 19B. The accuracy was quantified based on the RMSE between the desired location and the robot's final location, resulting in an average RMSE and standard deviation of 0.85±0.16 mm. It should be noted that the error is presumably lower for the MRI experiment due to the assumption that the center-line of the tube is in plane with the MRI's 2D slice, as shown in Fig. 19C.

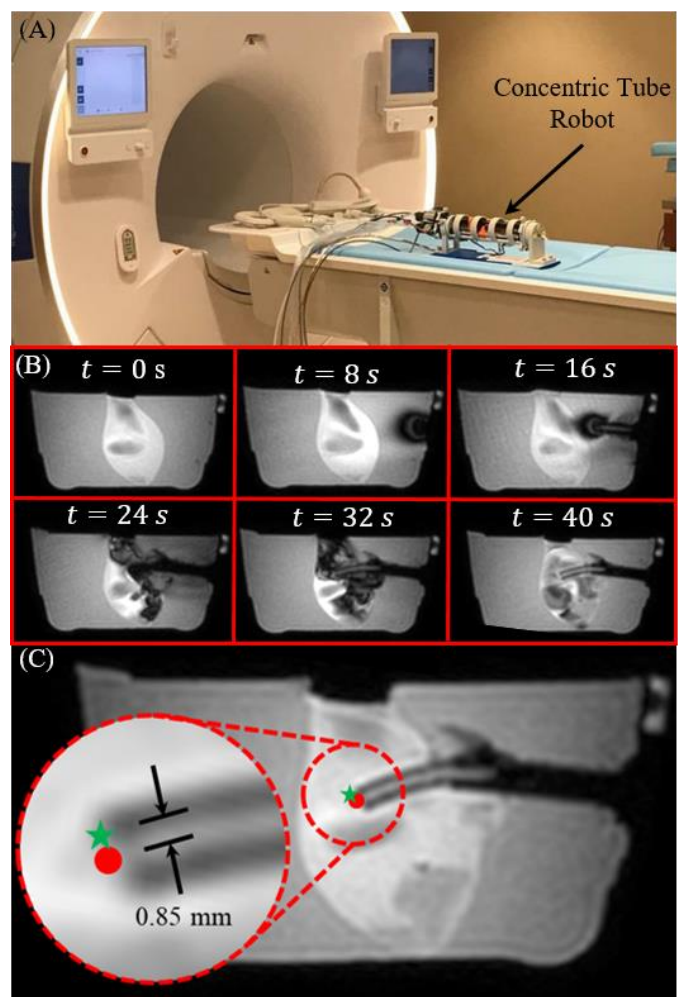

**Fig. 19.** (A) The MRI-based experimental setup can be seen. Note that the phantom and flexible coils are not depicted in the image. However, they would be placed in front of the robot. (B) A time sequence of the inner tube entering the phantom can be seen from start to finish in the sagittal plane MR images. Image intensity variations within the clot originate from artifacts caused by fluid flow during the evacuation. (C) An example of one dynamic slice can be seen with the difference between the target point (green star) and the resulting position (red dot) superimposed on the image. The clot material can be seen inside the aspiration tube.

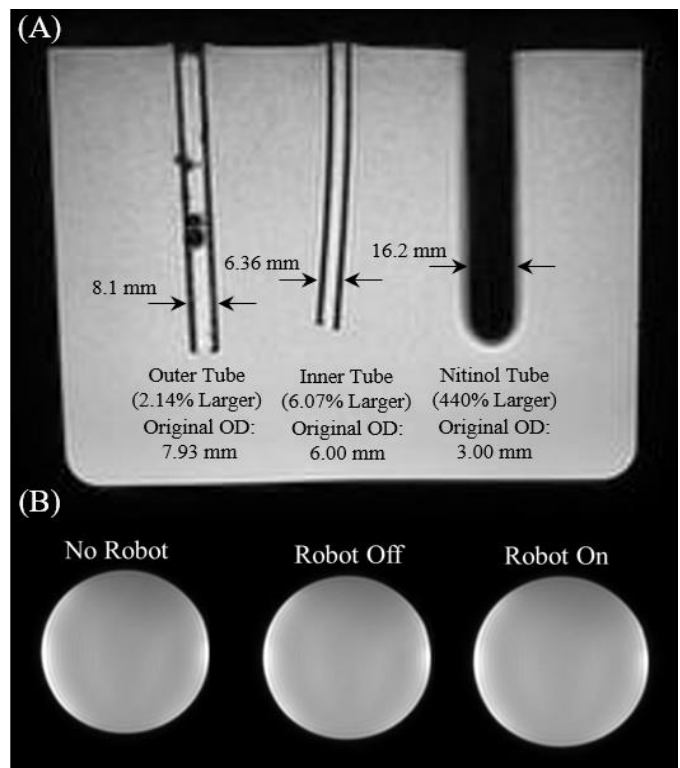

**Fig. 18.** (A) The outer tube, inner tube, and nitinol tube can be seen inside a water phantom in the coronal plane of the MRI. There was a percent increase of 2.14%, 6.07%, and 440% for the outer tube, inner tube, and nitinol tube, respectively. (B) $T_1$-weight images of the bottle phantom can be seen (left to right) with no robot, the robot turned off, and the robot turned on and operating.

### F. MR-Guided Phantom Clot Evacuation

The clot evacuation efficiency was first tested in a phantom trial using a 10% by weight Knox™ (Kraft Foods Global, Inc., USA) gelatin phantom. The phantom was filled with a congealed artificial clot substitute that provided high contrast during MR-imaging, resulting in a 38.36 mL clot (Fig. 20A-D). The experimental setup was identical to the MRI phantom accuracy characterization; however, target points were periodically selected to dynamically evacuate the clot, following the proposed workflow in Fig. 8.

After the procedure, the clot was segmented using 3D Slicer, an open-source and multi-platform software package widely used for medical applications [68]. The resulting segmentation can be seen in the lower sequence of Fig. 20. The initial volume was 38.36 mL and the remaining volume was 8.14 mL. Note that this is well under the 15 mL requirements discussed in the introduction section. The evacuation procedure lasted only 5 minutes.

## VII. DISCUSSION

The robotic system described in this article presents the first-ever use of off-the-shelf plastic tubes for CTR and exhibits promising results for their implementation for MRI-guided ICH evacuation. We have demonstrated the ability to characterize the shape of a tube with variable curvature and we were successfully able to extrapolate that to 3-D while considering torsional deflection when modeling the tube's shape, resulting in an average RMSE < 1.5 mm in both the free-space and MRI experiments. The robot was able to successfully evacuate a pseudo-clot in a phantom under dynamic MRI visualization to an acceptable residual hematoma volume while significantly reducing evacuation time compared to prior CTRs and state-of-the-art MIS approaches. The results can be seen in Table 3. Although these results indicate a promising future for MR-guided ICH evacuation, there is much work that needs to be done before this system can be implemented in a clinical setting

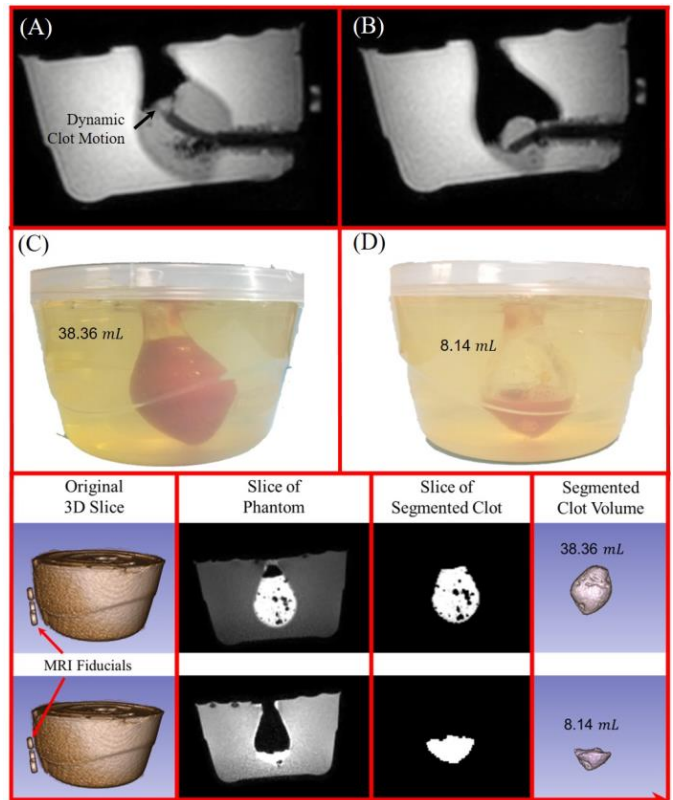

**Fig. 20.** (A) The tube can be seen within the phantom at the start of the evacuation procedure and (B) after the evacuation procedure. Note that (A) and (B) show the aspiration tube reaching two different target points during the procedure. (C) shows the initial clot volume and (D) shows the clot volume after evacuation. A 3D scan was performed before and after the procedure and the clot volume was segmented, as shown in the lower segmentation sequence.

– the most critical of which is dynamically updating the MR-image slice plane to enable closed loop aspiration monitoring and control. In our future work, we will sync the MRI workstation with the robot control GUI, allowing the MRI workstation to update the imaging plane dynamically based on the robot's joint space parameters, streamlining the clinician's interaction with the robot, and updating the treatment plan intraoperatively based on the image feedback.

Additionally, the robot hardware needs to be optimized in three different areas: (i) size and weight, (ii) workspace, and (iii) procedural lead-time. In regards to (i), the robot is currently longer than it needs to be and the mounting of the motors adversely contributes to this additional length. Our future work will focus on minimizing the robot's length and optimizing the mounting of the motors to further reduce the size and torque applied to the skull. There are two ways to engage in workspace optimization. (a) A one-size-fits-all CTM can be developed that

**Table 3.** Results of the robotic system

| Motor Accuracy [°] | TDoF[1] Accuracy [mm] | RDoF[2] Accuracy [°] | FK[3] Accuracy [mm] | IK[4] Accuracy [mm] | MRI Targeting Accuracy [mm] | Phantom Clot Evacuation [mL] |
|---|---|---|---|---|---|---|
| 0.317°±0.3° | 0.24±0.21 | 3.1±2.36 | 0.19±0.22 (Max: 2.04) | 1.39±0.54 (Max: 2.46) | 0.85±0.16 (Max: 1.15) | Initial: 38.36 Final: 8.14 Time: 5 minutes |

1. Translational Degree of Freedom accuracy, discussed in Section VI-C.
2. Rotational Degree of Freedom accuracy, discussed in Section VI-C.
3. Forward Kinematic accuracy of the tube tip and shape, discussed in Section VI-B.
4. Inverse Kinematic accuracy of the tube tip position with respect to the target, discussed in Section VI-C.

possesses optimized tube parameters based on a variety of clot dimensions and locations. (b) Tube dimensions are selected based on a patient by patient basis and the CTM that is optimal for the patient is used. Due to the ease in applying a desired curvature to our plastic tubes, our future work will focus on implementing (b). The primary bottle-neck in procedural lead-time is the manufacturing of the SF, causing a 4-hour delay. Although clinical trials assessing the benefit of hyperacute evacuation (within 6-12 hours of presentation) have yet to determine how significant the impact of a few hour delay may be, given the growing evidence that ICH is a time-sensitive pathology, this delay needs to be significantly reduced. Our future work will also investigate alternatives to a 3D printed SF, such as passively controlled SF legs, actively controlled SF legs, and modular attachments.

## VIII. CONCLUSIONS

In this article, we present the design and fabrication, mechanics modeling, and targeting validation of a CTR for MR-guided ICH evacuation. This article presents the first-ever use of off-the-shelf plastic tubes for CTR. Due to the variable curvature of these tubes and the low shear modulus of plastic, a new model was developed to analyze forward and inverse kinematics. This model used a single image of the inner tube at its maximum insertion displacement from the outer tube to characterize the curvature along the inner tube's centerline. The forward kinematics was then determined by iteratively applying the constant curvature transformation matrix along the tube's centerline. We then apply this model with the consideration of torsional deflection to determine inverse kinematics. Our model was able to obtain an RMSE of $1.39\pm0.54$ mm in a bench-top free-space setting and $0.85\pm0.16$ mm in our MRI phantom experiment. The evacuation efficiency of our robot was then tested in a phantom setting using a pseudo-clot. The robot was able to effectively evacuate the clot to an acceptable residual hematoma volume of 8.14 mL.

### ACKNOWLEDGMENT

The authors would like to thank Dylan Kingsbury and the rest of the team at Galil for their tireless assistance in implementing the Galil motion controller.